\pdfoutput=1

\documentclass[11pt]{article}

\usepackage{ACL2023}

\usepackage{times}
\usepackage{latexsym}
\usepackage{multirow}
\usepackage{amsmath}
\usepackage{array}
\usepackage{graphics}
\usepackage{pifont}
\usepackage{geometry}
\usepackage{arydshln}
\usepackage{graphicx}

\usepackage[T1]{fontenc}

\usepackage[utf8]{inputenc}

\usepackage{microtype}
\newcolumntype{P}[1]{>{\centering\arraybackslash}p{#1}}

\usepackage{inconsolata}

\title{Bidirectional Generative Framework for Cross-domain Aspect-based Sentiment Analysis}

\author{
\textbf{
Yue Deng \thanks{~~Yue Deng is under the Joint PhD Program between Alibaba and Nanyang Technological University.}~~\textsuperscript{\rm 1,2}~~
Wenxuan Zhang \thanks{$^\dag$ Wenxuan Zhang is the corresponding author.}$^\dag$\textsuperscript{\rm 1}~~ 
Sinno Jialin Pan\textsuperscript{\rm 2,3}}~~
Lidong Bing\textsuperscript{\rm 1}~~ \\
\textsuperscript{\rm 1}DAMO Academy, Alibaba Group~~
\textsuperscript{\rm 2} Nanyang Technological University, Singapore\\
\textsuperscript{\rm 3}Chinese University of Hong Kong~~
\\
{\tt\{yue.deng, saike.zwx, l.bing\}@alibaba-inc.com} \\
{\tt sinnopan@cuhk.edu.hk}
}

\begin{document}
\maketitle
\begin{abstract}
Cross-domain aspect-based sentiment analysis (ABSA) aims to perform various fine-grained sentiment analysis tasks on a target domain by transferring knowledge from a source domain. Since labeled data only exists in the source domain, a model is expected to bridge the domain gap for tackling cross-domain ABSA. Though domain adaptation methods have proven to be effective, most of them are based on a discriminative model, which needs to be specifically designed for different ABSA tasks. To offer a more general solution, we propose a unified bidirectional generative framework to tackle various cross-domain ABSA tasks. Specifically, our framework trains a generative model in both text-to-label and label-to-text directions.
The former transforms each task into a unified format to learn domain-agnostic features, and the latter generates natural sentences from noisy labels for data augmentation, with which a more accurate model can be trained. To investigate the effectiveness and generality of our framework, we conduct extensive experiments on four cross-domain ABSA tasks and present new state-of-the-art results on all tasks. Our data and code are publicly available at \url{https://github.com/DAMO-NLP-SG/BGCA}.
\end{abstract}

\section{Introduction}

Aspect-based sentiment analysis (ABSA) is the task of analyzing people's sentiments at the aspect level. 
It often involves several sentiment elements, including aspects, opinions, and sentiments \citep{sentimen-liu-2012, absasurvey-zhang-2022}.
For instance, given the sentence \textit{"The apple is sweet."}, the aspect is \textit{apple}, its opinion is \textit{sweet}, and the corresponding sentiment polarity is \textit{Positive}.
ABSA has attracted increasing attention in the last decade, and various tasks have been proposed to extract either single or multiple sentiment elements under different scenarios. For example, aspect sentiment classification (ASC) predicts the sentiment polarity of a given aspect target
\cite{chen-etal-2017-recurrent,li-etal-2018-transformation,xu-etal-2020-aspect} and
aspect term extraction (ATE) extracts aspects given the sentence \citep{DBLP:conf/ijcai/LiBLLY18,emnlp15-ate}, while aspect sentiment triplet extraction (ASTE) predicts all three elements in the triplet format \citep{aste-peng-2020, span-xu-2021}.

The main research line of ABSA focuses on solving various tasks within a specific domain.
However, in real-world applications, such as E-commerce websites, there often exist a wide variety of domains. 
Existing methods often struggle when applying models trained in one domain to unseen domains, due to the variability of aspect and opinion expressions across different domains \citep{hier-ding-2017, rnscn-wang-2018, memory-wang-2019}.
Moreover, manually labeling data for each domain can be costly and time-consuming, particularly for ABSA requiring fine-grained aspect-level annotation. 
This motivates the task of cross-domain ABSA, where only labeled data in the source domain is available and the knowledge is expected to be transferable to the target domain that only has unlabeled data. 

To enable effective cross-domain ABSA, domain adaptation techniques \citep{domain-adpat-Blitzer-2006, survey-sinno-2010} are employed to transfer learnt knowledge from the labeled source domain to the unlabeled target domain.
They either focus on learning domain-agnostic features \citep{hier-ding-2017, rnscn-wang-2018, adsal-li-2019}, or adapt the training distribution to the target domain \citep{uda-gong-2020, cdrg-yu-2021, generative-aope-li-2022-naccl}.
However, the majority of these works are based on discriminative models and need task-specific designs, making a cross-domain model designed for one ABSA task difficult to be extended for other tasks \citep{hier-ding-2017, rnscn-wang-2018, adsal-li-2019, uda-gong-2020}.
In addition, some methods further require external resources, such as domain-specific opinion lexicons \citep{cdrg-yu-2021}, or extra models for augmenting pseudo-labeled target domain data \citep{cdrg-yu-2021, generative-aope-li-2022-naccl}, which narrows their application scenarios. 

In a recent research line, pre-trained generative models like BART \citep{bart-lewis-2020} and T5 \citep{t5-raffel-2020} have demonstrated impressive power in unifying various ABSA tasks without any task-specific design and external resources.
By formulating each task as a sequence-to-sequence problem and producing the desired label words, \textit{i.e.}, the desired sentiment elements, they achieve substantial improvements on various ABSA tasks \citep{parapharase-zhang-2021, gas-zhang-2021, unifed-absa-yan-2021, seq2path-mao-2022-acl}.
Despite their success in supervised in-domain settings, their effectiveness has yet to be verified in the cross-domain setting.
Moreover, unlabeled data of the target domain, which is usually easy to collect, has shown to be of great importance for bringing in domain-specific knowledge \citep{survey-sinno-2010}. How to exploit such data with the generative formulation remains a challenge.  

Towards this end, we propose a \textbf{B}idirectional \textbf{G}enerative \textbf{C}ross-domain \textbf{A}BSA (BGCA) framework to fully exploit generative methods for various cross-domain ABSA tasks. 
BGCA employs a unified sequence-to-sequence format but contains two reverse directions: text-to-label and label-to-text. 
The text-to-label direction converts an ABSA task into a text generation problem, using the original sentence as input and a sequence of sentiment tuples as output.
After training on the source labeled data $\mathcal{D}^\mathcal{S}$, the model can then directly conduct inference on the unlabeled data $\mathbf{x}^\mathcal{T}$ of the target domain $\mathcal{D}^\mathcal{T}$ to get the prediction $\hat{\mathbf{y}}^\mathcal{T}$. 
The prediction can be used as pseudo-labeled data to continue-train the text-to-label model.
However, $\hat{\mathbf{y}}^\mathcal{T}$ is inevitably less accurate due to the domain gap between the source and target domains.
This is where the reverse direction, i.e., label-to-text, plays its role.

Specifically, we first reverse the order of input and output from the text-to-label stage of the source domain to train a label-to-text model. 
Then this model takes the prediction $\hat{\mathbf{y}}^\mathcal{T}$ as input and generates a coherent natural language text $\hat{\mathbf{x}}^\mathcal{T}$ that contains the label words of $\hat{\mathbf{y}}^\mathcal{T}$.
Note that even though the prediction $\hat{\mathbf{y}}^\mathcal{T}$ could be inaccurate regarding the original unlabeled data $\mathbf{x}^\mathcal{T}$, the generated sentence $\hat{\mathbf{x}}^\mathcal{T}$ can plausibly well match with $\hat{\mathbf{y}}^\mathcal{T}$. 
This is because the label-to-text model was trained to generate an output text that can appropriately describe the input labels.
Consequently, $\hat{\mathbf{y}}^\mathcal{T}$, drawn from the target domain, is able to introduce in-domain knowledge, thereby enhancing the overall understanding of the domain-specific information. 
In addition, $\hat{\mathbf{x}}^\mathcal{T}$ aligns more closely with $\hat{\mathbf{y}}^\mathcal{T}$ compared to $\mathbf{x}^\mathcal{T}$, which effectively minimizes the prediction noise. 
As such, they can be paired together to create a more accurate and reliable generated dataset.
Finally, the generated target data $\mathcal{D}^\mathcal{G}$ and the labeled source data $\mathcal{D}^\mathcal{S}$ can be combined to train the model in the text-to-label direction, which effectively enriches the model knowledge in the target domain. 

Our proposed BGCA framework exhibits some unique advantages. 
Firstly, it effectively utilizes the unlabeled target domain data by capturing important domain-specific words (i.e., sentiment elements) of the target domain in the first text-to-label stage.
In the meantime, it bypasses the issue from the domain gap since it takes the noisy prediction as input and obtains more accurate text-label pairs in the label-to-text stage.
Secondly, we fully leverage generative models' encoding and generating capabilities to predict labels and generate natural sentences within a unified framework, which is infeasible for discriminative models.
This allows the model to seamlessly switch between the roles of predictor and generator. 
Finally, BGCA utilizes a shared model to perform training in both directions, allowing for a more comprehensive understanding of the association between sentences and labels.

In summary, our main contributions are:
(1) We evaluate generative methods on four cross-domain ABSA tasks, including aspect term extraction (ATE), unified ABSA (UABSA), aspect opinion pair extraction (AOPE), and aspect sentiment triplet extraction (ASTE), and find that the generative approach is an effective solution. Without any unlabeled target domain data, it can already achieve better performance than previous discriminative methods.
(2) We propose a novel BGCA framework to effectively utilize unlabeled target domain data and train a shared model in reverse directions. 
It can provide high-quality augmented data
by generating coherent sentences given noisy labels and a unified solution to learn the association between sentences and labels thoroughly.
(3) Our proposed method achieves new state-of-the-art results on all tasks, which validate the effectiveness and generality of our framework.

\begin{figure*}[t]
\centering
\includegraphics[width=\linewidth]{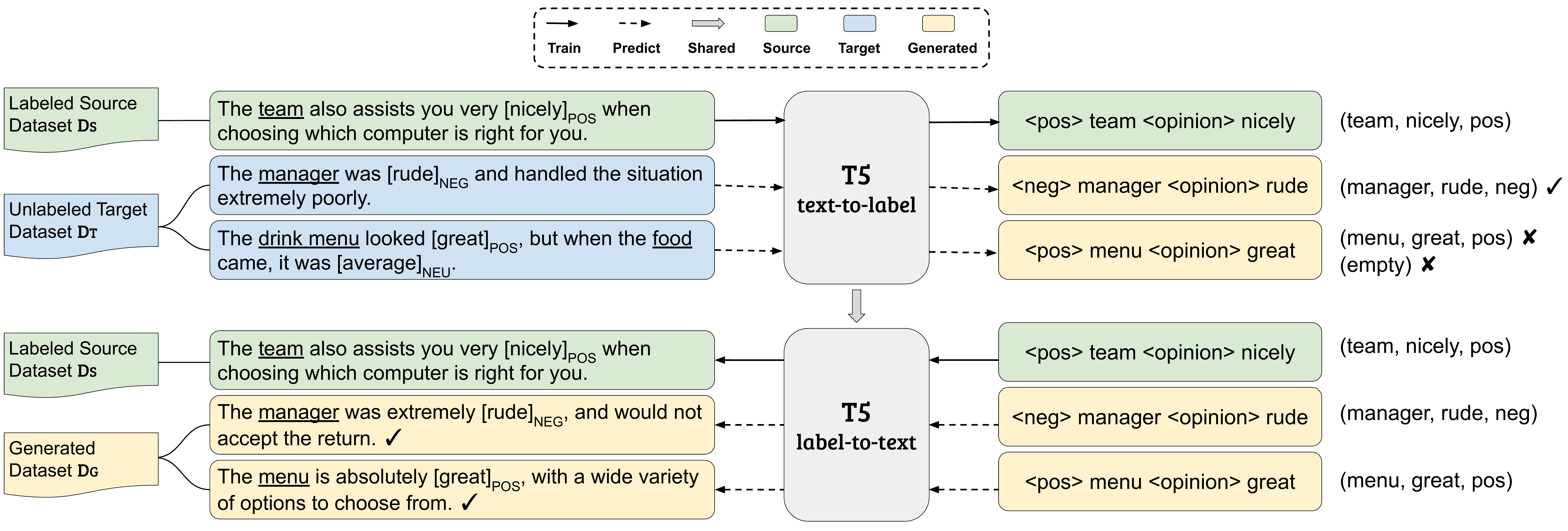}
\setlength{\belowcaptionskip}{-5pt}
\caption{
Overview of our proposed BGCA framework, which includes text-to-label and label-to-text directions. 
We take examples from the ASTE task for illustration.
Underlining and square brackets indicate gold aspects and gold opinions, respectively. 
The gold labels for the target domain are shown for demonstration only.
The generated dataset will be combined with the labeled source dataset to conduct final training in a text-to-label manner.
}
\label{fig:overview}
\end{figure*}

\section{Related Work}
\paragraph{Cross-domain ABSA}
Cross-domain ABSA aims to utilize labeled data from a source domain to gain knowledge that can be applied to a target domain where only unlabeled data is available.
The main research line of cross-domain ABSA involves two paradigms: feature-based adaptation and data-based adaptation \cite{absasurvey-zhang-2022}.
Feature-based adaptation focus on learning domain-invariant features.
Some have utilized domain-independent syntactic rules to minimize domain gap \citep{cross-jakob-2010, cross-chern-2014, hier-ding-2017, rnscn-wang-2018, memory-wang-2019},
while others have employed domain discriminators to encourage the learning of universal features \citep{adsal-li-2019, domain-classifier-Yang-2021, AHF-zhou-2021, domain-class-zhang-2023}.
On the other hand, data-based adaptation aims to adapt the training data distribution to the target domain. 
They either adjust the importance of individual training instances through re-weighting \citep{instance-xia-2014, uda-gong-2020}, or generate additional training data using another pre-trained model \citep{cdrg-yu-2021, generative-aope-li-2022-naccl}.
Despite their effectiveness, most of these works require task-specific design or external resources, preventing easy extensions to other cross-domain ABSA tasks. 

\paragraph{Generative ABSA}
Recently, generative models have obtained remarkable results in unifying various ABSA tasks.
By formulating each ABSA task as a sequence-to-sequence problem, generative models can output the desired sentiment element words \citep{gas-zhang-2021, seq2path-mao-2022-acl} or their indexes \citep{unifed-absa-yan-2021} directly.
In addition, some works successfully adopt the generative model on single ABSA tasks by converting the task to a natural language generation or paraphrase generation problem   \citep{generative-acs-liu-2021, parapharase-zhang-2021}.
Nevertheless, their potential is not explored under the cross-domain setting.

\section{Problem Formulation}

To examine the generality of our proposed framework, we consider four ABSA tasks, including ATE, UABSA, AOPE, and ASTE.
Given a sentence $\mathbf{x} = [w_1, w_2, ..., w_n]$ with $n$ words, the task is to predict a set of sentiment tuples denoted as $\mathbf{y} =\{t_i\}_{i=1}^{|t|}$,
where each tuple $t_i$ may include a single element from aspect ($a$), opinion ($o$), and sentiment ($s$), or multiple elements in pair or triplet format. The element within each tuple depends on the specific ABSA task, detailed in Table \ref{absa_task}. 

\begin{table}[t]
\centering
\scalebox{0.9}{
\begin{tabular}{c|c|c}
\hline
\textbf{Task} & \textbf{Output Tuple} & \textbf{Example Output}\\
\hline
ATE & $(a)$ & (apple)\\
UABSA & $(a, s)$ & (apple, positive)\\
AOPE & $(a, o)$ & (apple, sweet) \\
ASTE & $(a, o, s)$ & (apple, sweet, positive) \\
\hline
\end{tabular}
}
\setlength{\belowcaptionskip}{-5pt}
\caption{Output tuple of various ABSA tasks, and example output given the sentence \textit{"The apple is sweet."}, where $a$, $o$ and $s$ denote aspect, opinion and sentiment.}
\label{absa_task}
\end{table}

Under the cross-domain ABSA setting, the training dataset consists of a set of labeled sentences from a source domain $\mathcal{D}_{\mathcal{S}}=\left\{\mathbf{x}_i^{\mathcal{S}}, \mathbf{y}_i^{\mathcal{S}}\right\}_{i=1}^{N_{\mathcal{S}}}$ and a set of unlabeled sentences from a target domain $\mathcal{D}_{\mathcal{T}}=\{\mathbf{x}_j^{\mathcal{T}}\}_{j=1}^{N_{\mathcal{T}}}$. The goal is to leverage both $\mathcal{D}_{\mathcal{S}}$ and $\mathcal{D}_{\mathcal{T}}$ to train a model, which can predict the label of test data from the target domain. 

\section{Methodology}
We introduce our \textbf{B}idirectional \textbf{G}enerative \textbf{C}ross-domain \textbf{A}BSA (BGCA) framework in this section. 
As shown in Figure \ref{fig:overview}, it contains two sequential stages, namely text-to-label, and label-to-text, to obtain high-quality augmented data.
The text-to-label direction (on the top part) converts various tasks into a unified format and can produce noisy predictions on the unlabeled target data, whereas the label-to-text direction (on the bottom part) utilizes such noisy predictions to generate natural sentences containing the given labels so as to augment high-quality training data and enriches model knowledge of the target domain.

\subsection{Text-to-label} \label{sec:text-to-label}
The text-to-label direction unifies different ABSA tasks into a sequence-to-sequence format.
It takes a sentence as input and outputs a sequence of sentiment tuples extracted from the sentence.
We annotate the output sequence with predefined tagger tokens to ensure a valid format, which can prevent decoding ambiguity.
The tagger tokens are $k$ continuous tokens $\{\langle m_j \rangle\}_{j=1}^k$ initialized by embedding of the words $\{m_j\}_{j=1}^k$.
Specifically, we use $\langle aspect \rangle$, $ \langle opinion \rangle$ to mark aspect and opinion terms, and $\langle pos \rangle$, $ \langle neu \rangle$, $ \langle neg \rangle$ to annotate positive, neutral and negative sentiments.
The output formats with the continuous taggers for different tasks are:
\vspace{-0.3em}
\begin{equation}
\begin{array}{cl}
\text { ATE : } & \mathbf{x} \Rightarrow \langle aspect \rangle \, a \\
\text { UABSA : } & \mathbf{x} \Rightarrow \langle pos \rangle \, a \\
\text { AOPE : } & \mathbf{x} \Rightarrow  \langle aspect \rangle \, a \, \langle opinion \rangle \, o \\
\text { ASTE : } & \mathbf{x} \Rightarrow  \langle pos \rangle \, a \, \langle opinion \rangle \, o \\
\end{array}
\end{equation}
where $a$ and $o$ denote the aspect and the opinion terms, respectively. 
Taking ASTE as an example, we use the format of $\langle pos \rangle$ followed by the extracted aspect word(s), and $\langle opinion \rangle $ followed by the extracted opinion word(s) to annotate the positive opinion term expressed on the corresponding aspect term in a sentence.
Based on this format, we are able to extract the aspect, opinion, and sentiment from the output sequence to form a complete sentiment tuple through simple regular expressions.

The text-to-label direction is trained on $\{\mathbf{x}, \mathbf{y}\}$ pairs from $\mathcal{D}_{\mathcal{S}}$ by minimizing the standard maximum likelihood loss: 
\vspace{-0.3em}
\begin{equation}
\mathcal{L}=-\sum_{i=-1}^l \log p\left(y_i \mid \mathbf{x}; y_{\leq i-1}\right),
\end{equation}
where $l$ denotes the sequence length.

After training on the source labeled data $\mathcal{D}_{\mathcal{S}}$, we can directly conduct inference on the target domain  $\mathcal{D}_{\mathcal{T}}$ to extract the sentiment tuples $ \hat{\mathbf{y}}^\mathcal{T}$.
During the inference, we employ constrained decoding \citep{constrained-cao-2021} to ensure each generated token $\hat{y}^\mathcal{T}_i$ of the output sequence is selected from the input sentence or the predefined tagger tokens, in order to prevent invalid output sequences and ensure that the output is relevant to the specific domain:
\vspace{-0.3em}
\begin{equation}
\hat{y}_i^\mathcal{T} = \underset{y_j \in \mathcal{U}} {\operatorname{argmax}} \; p\left(y_j \mid \mathbf{x}^\mathcal{T} ;\hat{y}^\mathcal{T}_{\leq i-1}\right),
\end{equation}
where $\mathcal{U} =  \{w_i\}_{i=1}^n \cup \{\langle m_j \rangle\}_{j=1}^k$. 

\subsection{Label-to-text}
Although the text-to-label model can be directly applied for prediction on the target domain, it does not exploit the unlabeled target domain data in the training process, which has been proven to be crucial for incorporating target-domain knowledge \citep{survey-sinno-2010}.
One straightforward way to eliminate this problem is to use (${\mathbf{x}}^\mathcal{T}$, $\hat{\mathbf{y}}^\mathcal{T}$) as pseudo-labeled data to continue training the above text-to-label model.
However, such naive self-training suffers from the noise of $\hat{\mathbf{y}}^\mathcal{T}$.
Our label-to-text stage alleviates this weakness by pairing the label $\hat{\mathbf{y}}^\mathcal{T}$ with a new sentence that matches this label better.

Specifically, we continue to train the above model using the labeled dataset from $\mathcal{D}_{\mathcal{S}}$. 
Nevertheless, the training pairs are reversed into the label-to-text direction, where the input is now the sequence $\mathbf{y}$ with sentiment tuples, and the output is the original sentence $\mathbf{x}$:
\vspace{-0.3em}
\begin{equation}
\begin{array}{cl}
\text { ATE : } & \langle aspect \rangle \, a \Rightarrow \mathbf{x} \\
\text { UABSA : } & \langle pos \rangle \, a  \Rightarrow \mathbf{x} \\
\text { AOPE : } &  \langle aspect \rangle \, a \, \langle opinion \rangle \, o 
 \Rightarrow  \mathbf{x} \\
\text { ASTE : } &  \langle pos \rangle \, a \, \langle opinion \rangle \, o \Rightarrow  \mathbf{x} \\
\end{array}
\end{equation}

Similarly, the label-to-text direction is trained on $\{\mathbf{y}, \mathbf{x}\}$ pairs from $\mathcal{D}_{\mathcal{S}}$ by minimizing the standard maximum likelihood loss:
\vspace{-0.3em}
\begin{equation}
\mathcal{L}=-\sum_{i=-1}^{l'} \log p\left(x_i \mid \mathbf{y}; x_{\leq i-1}\right),
\end{equation}
and $l'$ refers to the sequence length.

\begin{table*}[ht]
\centering
\scalebox{0.9}{
\begin{tabular}{c|cccc|cccc|cccc}
\hline
\multirow{2}{*}{\textbf{Task}} & \multicolumn{4}{c|}{\textbf{ATE\&UABSA}} & \multicolumn{4}{c|}{\textbf{AOPE}} & \multicolumn{4}{c}{\textbf{ASTE}}\\ \cline{2-13} 
& L & R & D & S & L14 & R14 & R15 & R16 & L14 & R14 & R15 & R16 \\ \hline
Train & 3045 & 3877 & 2557 & 1492 & 1035 & 1462 & 678 & 971 & 906 & 1266 & 605 & 857 \\
Dev	& 304 & 387 & 255 & 149 & 116 & 163 & 76 & 108 & 219 & 310 & 148 & 210\\
Test & 800 & 2158 & 1279 & 747 & 343 & 500 & 325 & 328 & 328 & 492 & 322 & 326 \\
\hline
\end{tabular}
}
\caption{The statistics of ATE and UABSA}
\label{summary_statistics}
\end{table*}

After training, we use the sentiment tuples $ \hat{\mathbf{y}}^\mathcal{T}$, extracted from a target domain unlabeled data  ${\mathbf{x}}^\mathcal{T}$, to generate a natural sentence $\hat{\mathbf{x}}^\mathcal{T}$ incorporating the sentiment information in $ \hat{\mathbf{y}}^\mathcal{T}$.
To ensure fluency and naturalness, we decode the whole vocabulary set:
\vspace{-0.3em}
\begin{equation}
\hat{x}^\mathcal{T}_i = \underset{x_j \in \mathcal{V}} {\operatorname{argmax}} \; p\left(x_j \mid \hat{\mathbf{y}}^\mathcal{T} ;\hat{x}^\mathcal{T}_{\leq i-1}\right),
\end{equation}
where $\mathcal{V}$ denotes the vocabulary of the model.

The label-to-text stage thus augments a generated dataset $\mathcal{D}_{\mathcal{G}}=\left\{\hat{\mathbf{x}}^\mathcal{T}_i, \hat{\mathbf{y}}^\mathcal{T}_i\right\}_{i=1}^{N_\mathcal{T}}$.
By considering each natural sentence as a combination of context and sentiment elements, we can find that the generated sentence's context is produced by a model pre-trained on large-scale corpora and fine-tuned on the labeled source domain, while its sentiment elements such as aspects and opinions come from the target domain.
Therefore, $\mathcal{D}_{\mathcal{G}}$ can play the role of an intermediary which connects the source and target domains through the generated sentences.

As previously mentioned, due to the gap between source and target domains, the text-to-label model's prediction on unlabeled target data is noisy.
Instead of improving the text-to-label model, which may be difficult, our label-to-text stage creates a sentence $\hat{\mathbf{x}}^\mathcal{T}$ that is generated specifically for describing $\hat{\mathbf{y}}^\mathcal{T}$.
Thus, even with the presence of noise in the extracted labels $\hat{\mathbf{y}}^\mathcal{T}$, the label-to-text stage offers a means of minimizing the negative impact and ultimately yields a more accurate pseudo-training sample. 
Finally, since these two stages train a shared model based on sentences and labels from two directions, it gives the model a more comprehensive understanding of the association between sentences and labels, leading to a more accurate prediction of labels for given sentences.

\subsection{Training}
Ideally, the generated dataset $\mathcal{D}_{\mathcal{G}}$ should fulfil the following requirements: 
1) the natural sentence should exclusively contain sentiment elements that are labeled in the sentiment tuples, and should not include any additional sentiment elements;
2) the natural sentence should accurately convey all the sentiment elements as specified in the sentiment tuples without any omissions;
3) the sentiment tuples should be in a valid format and can be mapped back to the original labels;
Therefore, we post-process $\{\hat{\mathbf{x}}^t, \hat{\mathbf{y}}^t\}$  pairs from $\mathcal{D}_{\mathcal{G}}$ by:
1) filtering out pairs with $\hat{\mathbf{y}}^t$ in invalid format or contains words not present in $\hat{\mathbf{x}}^t$;
2) utilizing the text-to-label model to eliminate pairs where $\hat{\mathbf{y}}^t$ is different from the model's prediction on $\hat{\mathbf{x}}^t$.
In the end, we combine the source domain $\mathcal{D}_{\mathcal{S}}$, and the generated dataset $\mathcal{D}_{\mathcal{G}}$ as the ultimate training dataset and continue to train the same model in a text-to-label manner as outlined in Section \ref{sec:text-to-label}.

\section{Experiments}
\begin{table*}[th]
\centering
\scalebox{0.85}{
\begin{tabular}{l|ccc|ccc|cc|cc|c}
\hline
\hline
Methods & S→R & L→R & D→R & R→S & L→S & D→S &  R→L & S→L & R→D & S→D &\textbf{Avg.} \\ 
\hline
\hline
    \underline{\textit{\textbf{ATE}}} &  &  &  &  &  &  & & & & &\\ 
        $\text{Hier-Joint}^{\dagger}$ & 46.39 & 48.61 & 42.96 & 27.18 & 25.22 & 29.28 & 34.11 & 33.02 & 34.81 & 35.00 & 35.66 \\
        $\text{RNSCN}^{\dagger}$ &  48.89 & 52.19 & 50.39 & 30.41 & 31.21 & 35.50 & 47.23 & 34.03 & \textbf{46.16} & 32.41 & 40.84 \\
        $\text{AD-SAL}^{\dagger}$ &  52.05 & 56.12 & 51.55 & 39.02 & 38.26 & 36.11 & 45.01 & 35.99 & 43.76 & \textbf{41.21} & 43.91 \\
        $\text{BERT}_\text{B}\text{-UDA}^{\dagger}$ & 56.08 & 51.91 & 50.54 & 34.62 & 32.49 & 34.52 & 46.87 & 43.98 & 40.34 & 38.36 & 42.97 \\
        $\text{BERT}_\text{B}\text{-CDRG}^{\dagger}$ & 56.26 & 60.03 & 52.71 & 42.36 & 47.08 & 41.85 & 46.65 & 39.51 & 32.60 & 36.97 & 45.60 \\
        $\text{GAS}$ & 61.24 & 53.02 & 56.44 & 31.19 & 32.14 & 35.72 & 52.24 & 43.76 & 42.24 & 37.77 & 44.58 \\
        \hdashline
          $\text{BERT}_\text{E}\text{-UDA}^{\dagger*}$ & 59.07 & 55.24 & 56.40 & 34.21 & 30.68 & 38.25 & 54.00 & 44.25 & 42.40 & 40.83 & 45.53 \\
        $\text{BERT}_\text{E}\text{-CDRG}^{\dagger*}$ & 59.17 & 68.62 & 58.85 & \textbf{47.61} & \textbf{54.29} & 42.20 & 55.56 & 41.77 & 35.43 & 36.53 & 50.00 \\
        \hline 
        $\textbf{BGCA}_{\text{text-to-label}}$ & 60.03 & 55.39 & 55.83 & 36.02 & 35.43 & 37.73 & 54.18 & 43.45 & 42.49 & 37.89 & 45.84\\
        $\textbf{BGCA}_{\text{label-to-text}}$ & \textbf{63.20} & \textbf{69.53} & \textbf{65.33} & 45.86 & 44.85 & \textbf{54.07} & \textbf{57.13} & \textbf{46.15} & 37.15 & 38.24 & \textbf{52.15} \\
\hline
\hline
    \underline{\textit{\textbf{UABSA}}} &  &  &  &  &  &  & & & & &\\ 
        $\text{Hier-Joint}^{\dagger}$   & 31.10 & 33.54 & 32.87 & 15.56 & 13.90 & 19.04 & 20.72 & 22.65 & 24.53 & 23.24 & 23.72 \\
        $\text{RNSCN}^{\dagger}$   & 33.21 & 35.65 & 34.60 & 20.04 & 16.59 & 20.03 & 26.63 & 18.87 & 33.26 & 22.00 & 26.09 \\
        $\text{AD-SAL}^{\dagger}$   & 41.03 & 43.04 & 41.01 & 28.01 & 27.20 & 26.62 & 34.13 & 27.04 & 35.44 & 33.56 & 33.71 \\
        AHF & 46.55 & 43.49 & 44.57 & 33.23 & 33.05 & 34.96 & 34.89 & 29.01 & 37.33 & \textbf{39.61} & 37.67 \\
        $\text{BERT}_\text{B}\text{-UDA}^{\dagger}$  & 47.09 & 45.46 & 42.68 & 33.12 & 27.89 & 28.03 & 33.68 & 34.77 & 34.93 & 32.10 & 35.98 \\
        $\text{BERT}_\text{B}\text{-CDRG}^{\dagger}$  & 47.92 & 49.79 & 47.64 & 35.14 & 38.14 & 37.22 & 38.68 & 33.69 & 27.46 & 34.08 & 38.98 \\
        $\text{GAS}$ & 54.61 & 49.06 & 53.40 & 30.99 & 29.64 & 33.34 & 43.50 & 35.12 & 39.29 & 35.81 & 40.48 \\
        \hdashline
        $\text{BERT}_\text{E}\text{-UDA}^{\dagger*}$   & 53.97 & 49.52 & 51.84 & 30.67 & 27.78 & 34.41 & 43.95 & 35.76 & 40.35 & 38.05 & 40.63 \\
        $\text{BERT}_\text{E}\text{-CDRG}^{\dagger*}$  & 53.09 & 57.96 & 54.39 & 40.85 & \textbf{42.96} & 38.83 & \textbf{45.66} & 35.06 & 31.62 & 34.22 & 43.46 \\
        \hline
        $\textbf{BGCA}_{\text{text-to-label}}$ & 54.12 & 48.08 & 52.65 & 33.26 & 30.67 & 35.26 & 44.57 & 36.01 & \textbf{41.19} & 36.55 & 41.24 \\
        $\textbf{BGCA}_{\text{label-to-text}}$ & \textbf{56.39} & \textbf{61.69} & \textbf{59.12} & \textbf{43.20} & 39.76 & \textbf{47.94} & 45.52 & \textbf{36.40} & 34.16 & 36.57 & \textbf{46.07} \\
\hline
\hline
\end{tabular}
}
\caption{Results on cross-domain ATE and UABSA tasks. The best results are in bold. Results are the average F1 scores over 5 runs. $^\dagger$ denotes results from \citet{cdrg-yu-2021}, and the others are based on our implementation. $*$ represents methods that utilize external resources.}
\label{tab:ateuabsa}
\end{table*}

\subsection{Experimental Setup}

\begin{table*}[th]
\centering
\scalebox{0.85}{
\begin{tabular}{l|ccc|ccc|c}
\hline
\hline
Methods & R14→L14 & R15→L14 & R16→L14 & L14→R14 & L14→R15 & L14→R16 & \textbf{Avg.} \\ 
\hline
\hline
    \underline{\textit{\textbf{AOPE}}} &  &  &  &  &  &  &  \\ 
        $\text{SDRN}$ & 45.39 & 37.45 & 38.66 & 47.63 & 41.34 & 46.36 & 42.81 \\
        $\text{RoBMRC}$ & 52.36 & 46.44 & 43.61 & 54.70 & 48.68 & 55.97 & 50.29  \\
        $\text{SpanASTE}$ &  51.90 & 48.15 & 47.30 & 61.97 & 55.58 & 63.26 & 54.69 \\
        $\text{GAS}$ & 57.58 & 53.23 & 52.17 & 64.60 & 60.26 & 66.69 & 59.09 \\
    \hline
        $\textbf{BGCA}_{\text{text-to-label}}$ & 58.54 & 54.06 & 51.99 & 64.61 & 58.74 & 67.19 & 59.19\\
        $\textbf{BGCA}_{\text{label-to-text}}$ & \textbf{60.82} & \textbf{55.22} & \textbf{54.48} & \textbf{68.04} & \textbf{65.31} & \textbf{70.34} & \textbf{62.37} \\
\hline
\hline
    \underline{\textit{\textbf{ASTE}}} &  &  &  &  &  &  &  \\ 
        $\text{RoBMRC}$ & 43.90 & 40.19 & 37.81 & 57.13 & 45.62 & 52.05 & 46.12  \\
        $\text{SpanASTE}$ &  45.83 & 42.50 & 40.57 & 57.24 & 49.02 & 55.77 & 48.49 \\
        $\text{GAS}$ & 49.57 & 43.78 & 45.24 & 64.40 & 56.26 & 63.14 & 53.73 \\
        \hline
        $\textbf{BGCA}_{\text{text-to-label}}$ & 52.55 & \textbf{45.85} & 46.86 & 61.52 & 55.43 & 61.15 & 53.89 \\
        $\textbf{BGCA}_{\text{label-to-text}}$ & \textbf{53.64} & 45.69 & \textbf{47.28} & \textbf{65.27} & \textbf{58.95} & \textbf{64.00} & \textbf{55.80} \\
\hline
\hline
\end{tabular}
}
\setlength{\belowcaptionskip}{-8pt}
\caption{Results on cross-domain AOPE and ASTE tasks. The best results are in bold. Results are the average F1 scores over 5 runs. } 
\label{tab:aopeaste}
\end{table*}

\paragraph{Datasets}   
We evaluate the proposed framework on four cross-domain ABSA tasks, including ATE, UABSA, AOPE, and ASTE. 
Datasets of these tasks mainly consist of four different domains, which are Laptop (L), Restaurant (R), Device (D), and Service (S).
L, also referred to as L14, contains laptop reviews from SemEval ABSA challenge 2014 \citep{pontiki-etal-2014-semeval}.
R is a set of restaurant reviews based on SemEval ABSA challenges 2014, 2015, and 2016 \citep{pontiki-etal-2014-semeval, semeval15-task, pontiki-etal-2016-semeval}, denoted as R14, R15, R16 for the AOPE and ASTE tasks.
D contains digital device reviews provided by \citet{toprak-etal-2010-sentence}.
S includes reviews from web service, introduced by \citet{hu-2004-mining}.
Specifically, we can perform the ATE and UABSA tasks on all four domains, whereas the AOPE and ASTE tasks can be conducted on L and R domains, with R being further divided into R14, R15, and R16.
We follow the dataset setting provided by \citet{cdrg-yu-2021} for the ATE and UABSA task, and \citet{fan-etal-2019-target},  \citet{xu-etal-2020-position} for the AOPE, ASTE task respectively.
We show the statistics in Table \ref{summary_statistics}.

\paragraph{Settings}
We consider all possible transfers between each pair of domains for each task.
Following previous work \citep{Li_Bing_Li_Lam_2019, li-etal-2019-exploiting, uda-gong-2020, cdrg-yu-2021}, we remove D→L and L→D for the ATE and UABSA tasks due to their domain similarity. 
Additionally, we exclude transfer pairs between R14, R15, and R16 for the AOPE and ASTE tasks since they come from the same restaurant domain.
As a result, there are ten transfer pairs for the ATE and UABSA tasks, and six transfer pairs for the AOPE and ASTE tasks, detailed in Table \ref{tab:ateuabsa} and \ref{tab:aopeaste}.
We denote our proposed framework as $\textbf{BGCA}_{\text{label-to-text}}$, which includes the bidirectional augmentation and utilizes the augmented data for training the final model. 
To investigate the effectiveness of the generative framework for cross-domain ABSA tasks, we also report the results with a single text-to-label direction, denoted as $\textbf{BGCA}_{\text{text-to-label}}$, which is essentially a zero-shot cross-domain method.

\paragraph{Metrics}
We choose the Micro-F1 score as the evaluation metric for all tasks. 
A prediction is counted as correct if and only if all the predicted elements are exactly matched with gold labels.

\paragraph{Implementation Details}
We choose T5 \cite{t5-raffel-2020} as our backbone model and use T5-base checkpoint from \textit{huggingface}\footnote{\url{https://github.com/huggingface/}}. 
It is a transformer model \citep{attention-vaswani-2017} that utilizes the encoder-decoder architecture where all the pre-training tasks are 
in sequence-to-sequence format.
For simplicity, we use the Adam optimizer with a learning rate of 3e-4, a fixed batch size of 16, and a fixed gradient accumulation step of 2 for all tasks.
Regarding training epochs for text-to-label, label-to-text, and final training, we search within a range in \{15, 20, 25, 30\} using the validation set of the source domain for selection. 
We train our model on a single NVIDIA V100 GPU.

\subsection{Baselines}
For cross-domain ATE and UABSA tasks, we follow previous works to compare with established baselines including Hier-Joint \citep{hier-ding-2017}, RNSCN \citep{rnscn-wang-2018}, AD-SAL \citep{adsal-li-2019}, AHF \citep{AHF-zhou-2021}, $\text{BERT}_{\text{B/E}}$-UDA \citep{uda-gong-2020}, and $\text{BERT}_{\text{B/E}}$-CDRG \citep{cdrg-yu-2021} where $\text{BERT}_{\text{B}}$ and $\text{BERT}_{\text{E}}$ refer to models based on the original BERT and the continually trained BERT on large-scale E-commerce data containing around 3.8 million reviews \citep{berte-hu-2019}.
All of these methods utilize unlabeled target data, and $\text{BERT}_{\text{B/E}}$-CDRG are trained in a self-training manner,  which generates pseudo labels and retrain a new model with such labels.

For cross-domain AOPE and ASTE tasks, since there is no existing work on these two tasks under the cross-domain setting, we leverage the in-domain state-of-the-art models in a zero-shot manner for comparisons, including SDRN \citep{sdrn-chen-2020} for AOPE, and RoBMRC \citep{robmrc-liu-2022}, SpanASTE \citep{span-xu-2021} for ASTE task. 
In addition, we also refine RoBMRC and SpanASTE to work for the AOPE task by simply omitting the prediction of sentiment polarity. 

Most of the above baselines are discriminative methods based on the pre-trained BERT model. 
To enable a fair comparison, we also employ GAS \citep{gas-zhang-2021} for all four ABSA tasks, which is a strong unified generation method based on the same pre-trained generative model, i.e., T5-base, as our proposed BGCA method.

\begin{table}[t]
\centering
\small
\scalebox{0.87}{
\begin{tabular}{l|P{0.85cm}P{0.85cm}P{0.85cm}P{0.85cm}P{0.85cm}}
    \hline
    \textbf{Methods} & \textbf{ATE} & \textbf{UABSA} & \textbf{AOPE} & \textbf{ASTE} & \textbf{Avg.} \\ 
    \hline
    $\text{BGCA}^{\dagger}$ & \textbf{52.15} & \textbf{46.07} & \textbf{62.37} & 55.80 & \textbf{54.10} \\
     - self-training* & 46.13 & 41.56 & 61.33 & \textbf{55.99} & 51.25 \\
     - continue* & 46.63 & 42.22 & 58.56 & 54.70 & 50.53 \\
     - w/o sharing & 52.08 & 44.72 & 61.64  & 55.76 & 53.55 \\
    \hline
\end{tabular}
}
\setlength{\belowcaptionskip}{-10pt}
\caption{Ablation Study. $\text{BGCA}^{\dagger}$ represents our $\text{BGCA}_{\text{label-to-text}}$ setting. * denotes replacing the label-to-text stage with the corresponding training method.} 
\label{tab:ablation}
\end{table}

\begin{table*}[t]
    \centering
    \scalebox{0.8}{
    \begin{tabular}{m{6cm}|m{3.5cm}|m{6cm}}
    \hline
     \textbf{Sentence from R} &\textbf{Prediction}  &\textbf{Label-to-text Generation} \\
    \hline
     The [service$\text{]}_\texttt{POS}$ was good to excellent along with the [attitude$\text{]}_\texttt{POS}$. & (service, \texttt{POS}) & The [service$\text{]}_\texttt{POS}$  I received from Toshiba was excellent. \\
    \hline
    [Bottles of wine$\text{]}_\texttt{POS}$ are cheap and good. & (bottles, \texttt{POS}) &   I love the [bottles$\text{]}_\texttt{POS}$ they are made out of. \\
    \hline
     Our [waitress$\text{]}_\texttt{NEU}$ wasn't mean, but not especially warm or attentive either. & (waitress, \texttt{NEG}) &   The [waitress$\text{]}_\texttt{NEG}$ didn't even answer my question. \\
    \hline
    \end{tabular}
    }
    \setlength{\belowcaptionskip}{-10pt}    
    \caption{Examples on L→R from the UABSA task. Gold aspects are marked by square brackets. \texttt{POS}, \texttt{NEU} and \texttt{NEG} denote positive, neutral and negative sentiment. }
    \label{tab:case_study}
\end{table*}

\subsection{Main Results} 
We report the main results for the ATE and UABSA tasks in Table \ref{tab:ateuabsa} and the AOPE and ASTE tasks in Table \ref{tab:aopeaste}.
We have the following observations:
1) Our method with a single text-to-label direction ($\textbf{BGCA}_{\text{text-to-label}}$) establishes a strong baseline for cross-domain ABSA tasks.
Compared to discriminative baseline methods without external resources, it shows an improvement of 0.24\%, 2.26\%, 4.5\%, and 5.4\% on the cross-domain ATE, UABSA, AOPE, and ASTE tasks, respectively.
This demonstrates that generative models can actually generalize well across different domains with our designed continuous tagger to indicate the desired sentiment elements.
2) Our proposed framework $\textbf{BGCA}_{\text{label-to-text}}$ with bidiretional augmentations achieves new state-of-the-art results on all four cross-domain ABSA tasks.
It outperforms the previous best models by 2.15\% and 2.61\% on the ATE and UABSA tasks and by 3.28\% and 2.07\% on AOPE and ASTE.
Notably, it requires no external resources and can be seamlessly applied to all cross-domain ABSA tasks.
This verifies the generalizability and effectiveness of our proposed bidirectional generation-based augmentation method. 
3) Compared to other generation-based methods such as GAS and $\textbf{BGCA}_{\text{text-to-label}}$, $\textbf{BGCA}_{\text{label-to-text}}$ outperforms all of them on four tasks, indicating that the label-to-text direction can effectively utilize the unlabeled target data and leverage the potential of generative models.

\subsection{Ablation Study}
We conduct ablation studies to analyze the effectiveness of each component in BGCA. Results of different model variants are reported in Table \ref{tab:ablation}.

\paragraph{Ablation on label-to-text generation} 
To investigate the effectiveness of the label-to-text direction, and verify our assumption that it can fix the noisy prediction issue, 
we replace it with the self-training method and denote it as ``self-training'' in Table \ref{tab:ablation}.
Specifically, we use the pseudo labels of the unlabeled target domain data extracted by the text-to-label stage to replace our augmented data.
As shown in Table \ref{tab:ablation}, the performance drops about three points on average for four tasks.
This indicates that the pseudo-labeled samples from the text-to-label model contain more noise. 
Adding label-to-text generation could effectively address this issue by generating pseudo-training data with less noise.
To further investigate the effectiveness of generated samples, we manually check some samples on L→R from the UABSA task and show some representative samples in Table \ref{tab:case_study}.
Note that the gold labels for the target domain are not available during training, and we display them here for investigation only.
The first two example's predictions either omit an aspect or gives an incomplete aspect, while the third example's prediction gives the wrong sentiment.
However, the label-to-text model can generate a correct sentence that appropriately describes the prediction, although it is inaccurate regarding to the original input sentence.
These examples demonstrate how the label-to-text stage can resolve noisy prediction issues and produce high-quality target domain data.

\begin{figure}[t]
\centering
\includegraphics[scale=0.35]{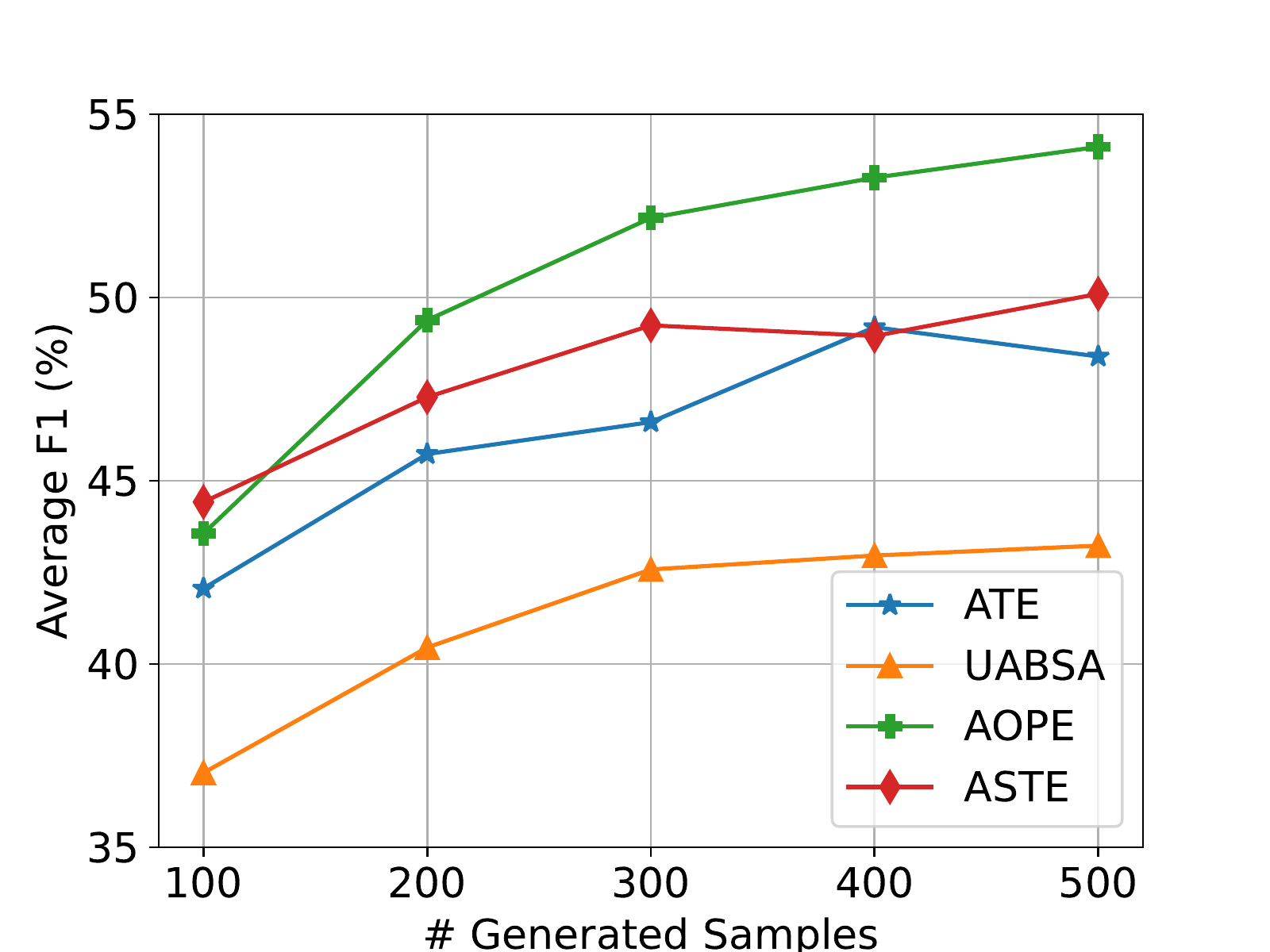}
\setlength{\belowcaptionskip}{-10pt}
\caption{Comparison results of our method with a different number of generations. 
}
\label{fig:aug_num}
\end{figure}

\paragraph{Ablation on unlabeled data utilization}
Continue training has shown to be an effective method to leverage unlabeled data by conducting pre-training tasks on relevant corpora to capture domain-specific knowledge \citep{berte-hu-2019, uda-gong-2020, cdrg-yu-2021}.
We compare it with our method to discuss how to utilize unlabeled data for generative cross-domain ABSA and denote it as ``continue'' in Table \ref{tab:ablation}.
Specifically, we replace the label-to-text stage with conducting continue-train on the unlabeled data of the target domain, with the span reconstruction objective as original T5 pre-training \citep{t5-raffel-2020}.
The results show that continue training lies behind our proposed method and demonstrate that our framework can effectively utilize unlabeled target domain data. 
The possible reason may be that continue training requires many training samples, which is infeasible in cross-domain ABSA scenarios.

\paragraph{Ablation on model sharing}
To demonstrate the advantages of training a shared model in both directions, we compare it to a method where a model is newly initialized before each stage of training and denote it as ``w/o sharing'' in Table \ref{tab:ablation}.  
Results on four tasks show that our approach outperforms the non-shared method by an average of 0.6\%, suggesting that a shared model owns a better understanding of the association between sentences and labels.

\subsection{Further Analysis}
\paragraph{Analysis on number of generated samples}
Figure \ref{fig:aug_num} shows the comparison results over four tasks with different numbers of generated samples.
To better analyze the effect of the number of generations, we exclude the source training data and solely use the generated samples for final training.
There is an apparent trend of performance improvement with the increasing number of generated samples, revealing that the generated samples can boost cross-domain ability.

\begin{table}[t]
\centering
\resizebox{\linewidth}{!}{
\begin{tabular}{c|cc|cc}
\hline
 \multirow{2}{*}{\textbf{Group}} & \multicolumn{2}{c|}{\textbf{ATE}} & \multicolumn{2}{c}{\textbf{UABSA}} \\ \cline{2-5} 
& text→label & label→text & text→label & label→text \\ \hline
Zero & 45.31 & 36.48 & 50.02 & 39.18 \\
Single & 41.53 & 47.99 & 35.02 & 43.17 \\
Multiple & 26.61 & 37.20 & 21.99 & 29.59 \\ \hline
\end{tabular}
}
\caption{Comparison results on cross-domain ATE and UABSA tasks over different sentence groups containing zero, single, or multiple aspects respectively.}
\label{tab:aspect_num}
\end{table}

\paragraph{Analysis on improvement types}
To understand what types of cases our method improved, we categorize sentences from the test set into three groups: without any aspect, with a single aspect, and with multiple aspects.
We conduct our analysis on the cross-domain ATE and UABSA tasks since they contain sentences without any aspect, and evaluate the performance of both the text-to-label and label-to-text settings for each group.
We choose sentence-level accuracy as the evaluation metric, \textit{i.e.}, a sentence is counted as correct if and only if all of its sentiment elements are correctly predicted.
We present the average accuracy across all transfer pairs in Table \ref{tab:aspect_num}.
The text-to-label model has less knowledge of the target domain and thus tends to predict sentences as no aspect, leading to high accuracy in the group without any aspect.
However, it also misses many sentiment elements in the other two groups.
On the other hand, although label-to-text lies behind text-to-label in the group without any aspect, it significantly improves the performance of sentences with single or multiple aspects.
This indicates that the label-to-text model has obtained more target domain knowledge than the text-to-label setting, and thus can identify more sentiment elements. 

\section{Conclusions}
In this work, we extend the generative method to cross-domain ABSA tasks and propose a novel BGCA framework to boost the generative model's cross-domain ability.
Specifically, we train a shared generative model in reverse directions, allowing high-quality target domain augmentation and a unified solution to comprehend sentences and labels fully.
Experiments on four cross-domain ABSA tasks verify the effectiveness of our method.

\section{Limitations}
In this paper, we present a bidirectional generative framework for cross-domain ABSA that has achieved outstanding results on four cross-domain ABSA tasks. 
Although there is only one stage during inference, our method involves multiple training stages, including text-to-label, label-to-text, and final training. 
These additional training stages not only lengthen the training time but also require additional computational resources, which may hinder scalability for large-scale data and result in a burden for the environment.

\section*{Acknowledgements}
S. J. Pan thanks for the support of the Hong Kong Global STEM Professorship. Y. Deng is supported by Alibaba Group through Alibaba Innovative Research (AIR) Program and Alibaba-NTU Singapore Joint Research Institute (JRI), Nanyang Technological University, Singapore. 

\bibliography{custom}
\bibliographystyle{acl_natbib}

\end{document}